\documentclass[10pt, a4paper]{article}
\usepackage{lrec}
\usepackage{graphicx}
\usepackage{tabularx}
\usepackage{soul,xcolor}

\usepackage[utf8]{inputenc}
\usepackage[english]{babel}

\usepackage{epstopdf}
\usepackage{hyperref}
\usepackage{xstring}

\usepackage{color,graphicx}
\usepackage{tikz}
\usepackage{tikz-dependency}
\usetikzlibrary{shapes, arrows, positioning, calc, automata, decorations.pathreplacing}
\usepackage{wrapfig}
\definecolor{pf1}{RGB}{158, 81, 5}
\definecolor{pf2}{RGB}{29, 143, 166}
\usepackage{ulem}
\usepackage[bottom]{footmisc}

\newcommand\redout{\bgroup\markoverwith {\textcolor{pf1}{\rule[0.40ex]{2pt}{0.65pt}}}\ULon}

\title{LSCP: Enhanced Large Scale Colloquial Persian Language Understanding}

\name{{Hadi Abdi Khojasteh\textsuperscript{1}}, {\textbf{Ebrahim Ansari} \textsuperscript{1,2}}, {\textbf{Mahdi Bohlouli} \textsuperscript{1,3}}}

\address{\textsuperscript{1} Department of Computer Science and Information Technology, \\
Institute for Advanced Studies in Basic Sciences (IASBS), Zanjan, Iran \\
\textsuperscript{2} Institute of Formal and Applied Linguistics (ÚFAL), \\
Faculty of Mathematics and Physics, Charles University in Prague, Czechia \\
\textsuperscript{3} Research and Innovation Department, Petanux GmbH, Germany \\
         \{hkhojasteh, ansari, bohlouli\}@iasbs.ac.ir\\}
         
\abstract{
Language recognition has been significantly advanced in recent years by means of modern machine learning methods such as deep learning and benchmarks with rich annotations. However, research is still limited in low-resource formal languages. This consists of a significant gap in describing the colloquial language especially for low-resourced ones such as Persian. In order to target this gap for low resource languages, we propose a ``Large Scale Colloquial Persian Dataset" (LSCP). LSCP is hierarchically organized in a semantic taxonomy that focuses on multi-task informal Persian language understanding as a comprehensive problem. This encompasses the recognition of multiple semantic aspects in the human-level sentences, which naturally captures from the real-world sentences. We believe that further investigations and processing, as well as the application of novel algorithms and methods, can strengthen enriching computerized understanding and processing of low resource languages. The proposed corpus consists of 120M sentences resulted from 27M tweets annotated with parsing tree, part-of-speech tags, sentiment polarity and translation in five different languages.
\\ \newline  }

\begin{document}

\maketitleabstract

\section{Introduction}
Language understanding is a comprehensive problem that encompasses the recognition of multiple semantic aspects of relations between different linguistic units and compounds: homonymy, synonymy, antonymy, hypernymy, hyponymy, meronymy, metonymy, holonymy, paronyms. In other hands, large scale corpora are one of the key resources in Natural Language Processing (NLP). In spite of their importance in many lingual applications, no large-scale Persian corpus has been made available so far, given the difficulties in its creation and the intensive labour required due to its low resources, especially in colloquialism.

Today, modern machine learning algorithms such as deep neural networks provide improved processing performance for a wide variety of applications, mainly text mining and NLP. But, they need large scale datasets, specially for training and pre-processing of the algorithms. It should also be mentioned that collecting, preparing, pre-processing and annotating large scale datasets from social media and web based materials such as product reviews are not easy task. Because most of them are associated with the mixed multilingual and informal texts that need better understanding. Social networks such as Twitter deliver large volume of disparate data including short and long mixed and informal text consisting symbols, other languages and images.

\begin{figure}\begin{center}
  \includegraphics[scale=0.94]{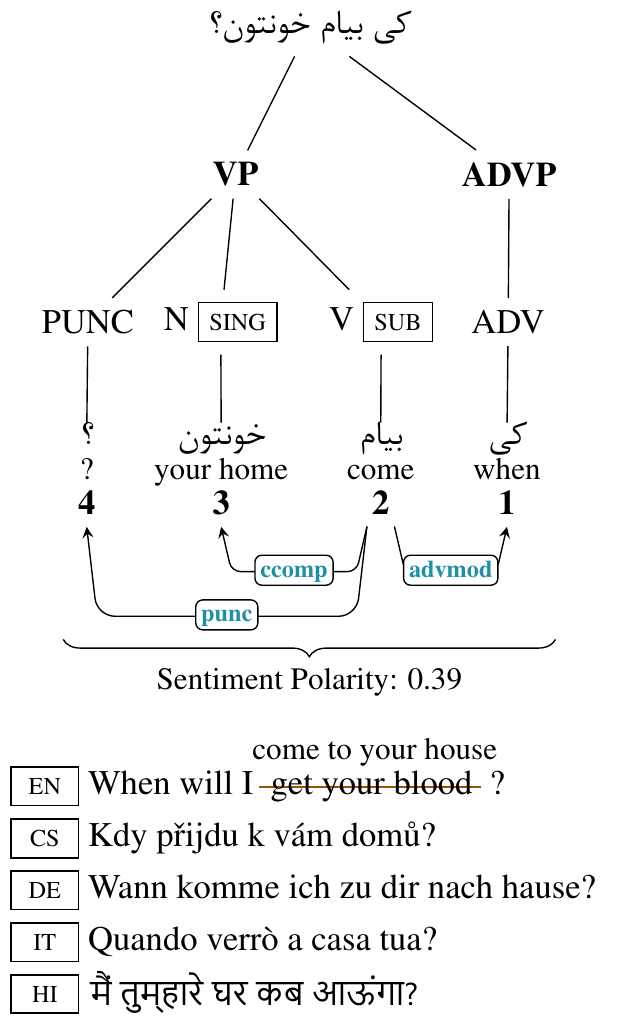} 
  \caption{A sample of the extracted sentence with dependency relations in syntactic annotation, part-of-speech tags, sentiment polarity and translations.}
  \label{fig:LSCP single sample}
\end{center}\end{figure}

The quality of machine translation depends heavily on availability of various datasets. In this regard, low resourced languages such as Persian lack the quality of machine translation due to missing enriched datasets. It should be stated that releasing enhanced dataset for low resourced languages can also support the use of new methods such as Neural Machine Translation (NMT) in these languages in order to improve their performance and accuracy and also enhance rule-based and statistical methods. 

In the frame of this research, we propose a new dataset for Large-Scale Colloquial Persian (LSCP) language understanding. The LSCP contains approx. 120M sentences from 27M tweets in total with annotations for training and validation set spanning over words and sentences. LSCP enriches semantic aspects of daily events, news, sports and politics, which naturally captures from the real-world sentences. Furthermore, we propose a new temporal deep neural network architecture called Colloquialism Temporal Network (CATNet) that builds on top of recurrent architecture by mid-level representations and temporal cues. CATNet focuses on multi-task learning and is trained in an end-to-end manner. The experiments show that LSCP enhances current state-of-the-art methods on formal language datasets, especially multi-domain neural machine translation. The dataset is publicly available in the LINDAT/CLARIN repository \cite{lindatData}.

\begin{table}
	\begin{center}
		\begin{tabular}{c c c} 
		    \hline
			Train & Validation & Test \\ \hline
			22.5M & 1.5M & 3M \\ \hline
		\end{tabular}
	\end{center}
	\caption{Dataset statistics for train, validation, and test sets.}
	\label{table:LSCP statistics splits}
\end{table}

\subsection{Towards the Persian Colloquial Language}
The language in oral form is typically much more dynamic than its written. The written variety of a language typically involves a higher level of ritual, whereas the spoken form is characterised by several contractions and abbreviations. In formal written texts, longer and tougher sentences tend to be used, as a result of the reader can re-read the troublesome parts, if they lose track. The spoken form is shorter, additionally thanks to semantic augmentation by visual cues that don't seem to be available in the written forms. The size of vocabulary in use is one of the foremost noticeable variations between oral and written sorts of discourse. The written language uses synonyms rather than continuance an equivalent word over and all over again. This is, however, not the case in oral language, which typically uses a lot of restricted vocabulary. The extent of difficulty in pronunciation may additionally have an effect on the words chosen. Oral languages tend to use words of fewer syllables.

In addition to aforesaid general differences between spoken and written varieties of languages, the Persian language introduces range of variations that further expand this gap. Additionally, several informal words are not applicable to be used in formal language, so that there are outstanding variations in pronunciation of words.
This alteration is quite common, but has no rule. The speaker is not allowed to interchange various words in the colloquial language. Obviously, Persian tweets involve all of earlier stated features of spoken form of it, which is rare in formal writings.

\section{Related Work}
Nowadays, almost all textual data for NLP are based on social media and online sources such as retrieved ones from the web. 
The social media generated data is, in fact, informal, multilingual and mixed, which needs complex methods to understand them. This problem becomes even more difficult, when it comes to low resource languages such as Persian. In this regard, we review the literature first for informal language understanding and then concentrate on Persian language datasets. 

\subsection{Informal Language Understanding Datasets}
\newcite{lo2017multilingual} argued that the social media data is typically multilingual and it is mostly difficult to grasp the full sentiment of online user-generated product content specially from social networks. They already used user-generated product reviews as an informal language datasets and also focused on the mixed linguistic origin of online social media languages and accordingly reviewed existing tools and state-of-the-art research on sentiment analysis works and also provided recommendations in this regard. \newcite{zhao2018lsicc}  proposed a large scale informal Chinese Corpus. This dataset consists of 37M book reviews and 50 thousand netizen’s comments. 

\subsection{Persian Language Understanding Datasets}
\subsubsection{Part-of-Speech Tagger}
Bijankhan corpus \cite{bijankhan2004role} is a tagged corpus that is suitable for natural language processing research on the Persian language. This collection is gathered form daily news and common texts. In this collection all documents are categorized into different subjects such as political, cultural and so on. Totally, there are 4300 different subjects. The Bijankhan collection contains about 2.6M manually tagged words with a tag set that contains 40 Persian POS tags.
FarsNet \cite{shamsfard2010semi} is a lexical ontology for the Persian language that is designed to contain a Persian WordNet with about more than 30K entries organized in about 20K synsets and verbs argument structures with their selectional restrictions. 
MULTEXT-East \cite{erjavec2012multext} is a multilingual dataset for language engineering research and development. This dataset contains word-forms, lemmas and MSDs for about 2M words.
Uppsala Persian Dependency Treebank (UPDT) \cite{seraji2015morphosyntactic} is a dependency-based syntactically an notated corpus.
Persian Morphologically Segmented Lexicon \cite{ansari2019persian} dataset includes 45300 Persian word forms which are manually segmented into sequences of morphemes.

\subsubsection{Named Entity Recognition and Dependency Parsing}
Persian Syntactic Dependency Treebank \cite{rasooli2013development} has 29,982 annotated sentences including samples from almost all verbs of the Persian valency lexicon.
ArmanPersoNERCorpus \cite{hafezi2018neural} includes 250K tokens and 7K Persian sentences in total and is available in 3 folds to be used in turn as training and test sets contains tokens, along with its manually annotated named-entity and Named Entity Recognition tags.
Universal Dependencies \cite{nivre2017universal} is a cross-linguistically consistent corpus that holds, with the goal of facilitating multilingual parser development, cross-lingual learning, and parsing research from a language typology perspective.

\subsubsection{Text Categorization and Spell Checking}
FASpell \cite{barari2005clonizer} dataset was developed for the evaluation of spell checking algorithms. It contains a set of pairs of misspelled Persian words and their corresponding corrected forms similar to the ASpell dataset used for English.
Hamshahri collection \cite{aleahmad2009hamshahri} is a standard reliable Persian text collection that was used at Cross Language Evaluation Forum (CLEF) for evaluation of Persian information retrieval systems.

\subsubsection{Machine Translation}
OPUS \cite{tiedemann2012parallel} is a growing collection of translated texts from the web. In the OPUS project try to convert and align free online data, to add linguistic annotation, and to provide the community with a publicly available parallel corpus.
English-Persian Parallel Corpus \cite{karimi2017extracting} consists of about 200K bilingual paired sentences which have been sorted by their degree of similarity.
Tehran English-Persian Parallel (TEP) Corpus \cite{pilevar2011tep} is a English-Persian corpus constructed using movie subtitles from 21K files.
W2C \cite{majlivs2011w2c} corpora is a set of corpora for 120 languages automatically collected from Wikipedia and the web.
dotIR \cite{darrudi2008dorir} is a standard Persian test collection that is suitable for evaluation of web information retrieval algorithms contains Persian web pages including their text, links and metadata. It is A good test bed for evaluation of link based information retrieval algorithms and also includes enough queries for a valid evaluation.

\section{LSCP Dataset}
The Large Scale Colloquial Persian Language Understanding dataset (LSCP) is organized hierarchically in a semantic taxonomy of colloquialism understanding. Almost, all real-wold conditioned Persian datasets are targeting semantic analytics of formal language. However, a language is not only formal, which provides a clear, properly framed and well organised. By focusing on formal writing, we ignore the information about abbreviations, unspecified vocabulary, contractions, jargons, slang, cliche, colloquial diction and ordinary sentences. To our knowledge, there is no colloquial large scale corpus publicly available. The LSCP has 120M sentences from 27M casual Persian sentences with its derivation tree, part-of-speech tags, sentiment polarity and translations in English, German, Czech, Italian and Hindi spoken languages. Figure~\ref{fig:LSCP single sample} shows a sample of the extracted dataset with its annotation.
One of the important research questions which is not addressed well in recent works on language learning, is leveraging the informal language information in sentences. The LSCP dataset makes it possible to assess the effect of learning and knowledge transfer among different tasks, such as enabling transfer learning of language in formal to informal and vice-versa or low-resource cross-lingual transfer. In summary, the LSCP can help the NLP community and bring more interesting solutions to colloquial language.

\begin{figure}\begin{center}
  \includegraphics[scale=1.0]{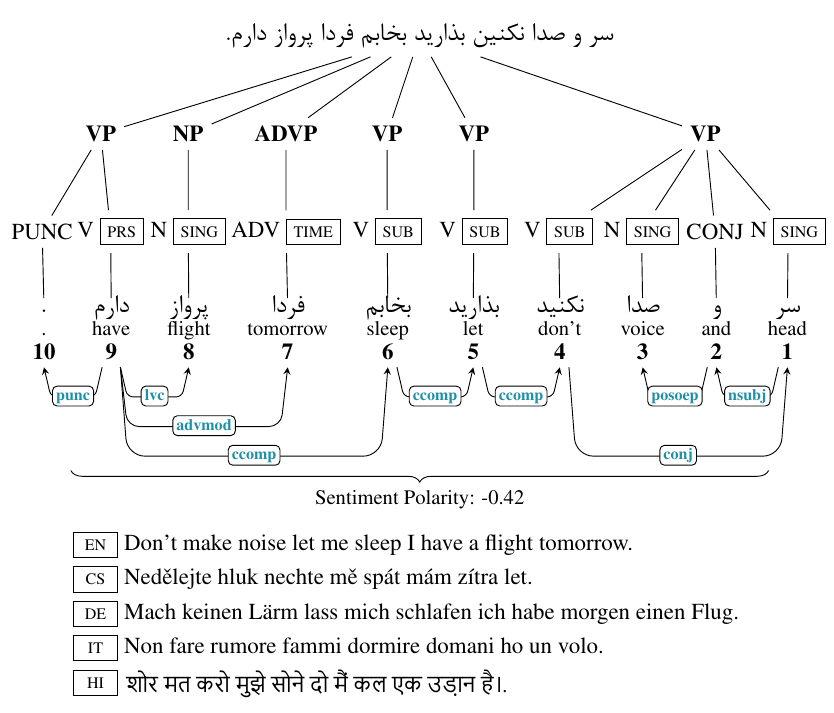} 
  \caption{A sample of the extracted sentence with dependency relations in syntactic annotation, part-of-speech tags, sentiment polarity and translations.}
\label{fig:LSCP second sample}
\end{center}\end{figure}

\subsection{LSCP Statistics}
The LSCP consists of \textbf{27M} tweets. The number of samples for train, validation, and test splits are reported in Table~\ref{table:LSCP statistics splits}. The dataset consists of usual sentences. In practice, the length of the sentences are different with a maximum of 40 words length. LSCP contains human language texts into lists of sentences and words, generated base forms of those words, their parts of speech (Table~\ref{table:UPCPOS}), a syntactic structure dependency parse (Table~\ref{table:SRUPDT}) for each word and translation in five languages as it is shown in Figure~\ref{fig:LSCP second sample}.

\begin{table}\begin{center}\resizebox{\columnwidth}{!}{
\renewcommand{\arraystretch}{0.9}
\begin{tabular}{ l | l | l | l } 
 \cline{1-2}
 \multicolumn{1}{l |}{Category} & \multicolumn{2}{l}{Description} &  \\
 \hline \hline
 ADJ & Adjective & INT & Interjection\\
 ADJ\_CMPR & {\small Comparative adjective} & N\_PL & Plural noun\\
 ADJ\_INO & Participle adjective & N\_SING & Singular noun\\
 ADJ\_SUP & Superlative adjective & NUM & Numeral\\
 ADJ\_VOC & Vocative adjective & N\_VOC & Vocative noun\\
 ADV & Adverb & P & Preposition\\
 ADV\_COMP & {\small Adverb of comparison} & PREV & {\small Preverbal particle}\\
 ADV\_I & {\small Adverb of interrogation} & PRO & Pronoun\\
 ADV\_LOC & {\small Adverb of location} & SYM & Symbol\\
 ADV\_NEG & {\small Adverb of negation} & V\_AUX & Auxiliary verb\\
 ADV\_TIME & Adverb of time & V\_COP & Verb copula\\
 CLITIC & Accusative marker & V\_IMP & Imperative verb\\
 CON & Conjunction & V\_PA & Past tense verb\\
 DELM & Delimiter & V\_PP & {\small Past participle verb}\\
 DET & Determiner & V\_PRS & {\small Present tense verb}\\
 FW & Foreign Word & V\_SUB & Subjunctive verb\\
 \cline{1-4}
\end{tabular}}\end{center}
\caption{Part-of-speech tags in UPC.}
\label{table:UPCPOS}
\end{table}

\subsection{Collecting and Annotation}
Building a large-scale language understanding dataset is a time-consuming task. In practice, there are two main tasks, which are usually most time consuming for creating a large-scale corpus: (a) data collection and (b) annotation.
We collected 27M tweets by a crawler  \cite{khojasteh2020deep} based on the Twitter API. To make sure that the collection is diverse, we first created a list of seed users. Then, the followers of the seed accounts were added to the list. Next, the latest tweets of the users in our list were extracted and saved in the dataset. To make the data appropriate for our task, we removed the same tweets, non-Persian tweets and the ones that had fewer words. This led the dataset to have relatively long sentences with diverse concept.

For the annotation of the datasets, we adopt a semi-automatic crowd-sourcing strategy, in which a human manually verifies the crawled sentences, to reduce the cost of data collection and annotation. Manually annotating a large number of sentences with multiple semantic categories (i.e thousands of concepts and tags) has two major shortcomings, (a) manual annotations are error-prone because a human cannot be attentive to every detail occurring in the context that leads to mislabeling and are difficult to eradicate; (b) large scale sentence annotation in specific is very time consuming task due to the amount and various sort of tags. To overcome these issues, we employ a two-stage framework for the LSCP annotation. In the first stage, we utilize the StanfordNLP \cite{qi2018universal} to get rough annotations of the sentences. The model predicts tags per tweet. Annotations consist lists of sentences and words, base forms of those words, their parts of speech, and a syntactic structure dependency.

For translation, we utilized Google Cloud Translation\footnote{https://cloud.google.com/translate} to provide translations. Then, generated sentences are compared to the output of the combination of Moses \cite{koehn2007moses} and Transformer \cite{vaswani2017attention} model implemented with \cite{klein2017opennmt} and trainned on XNLI \cite{conneau2018xnli} with fine-tunning on English-Persian Parallel Corpus \cite{karimi2017extracting}.

In the second stage, we apply human verification to remove any possible mislabeled noisy tags and also add possible missing tags by the model from recommended list. They were also able to edit words or phrases in English translation for improving the overall quality of parallel sentences. Figure~\ref{fig:LSCP second sample} shows sample of the extracted dataset with its annotation. Final dataset is enriched with 64M labels for words and over 130M translated sentences.

\begin{table}\begin{center}\resizebox{0.35\textwidth}{!}{
\renewcommand{\arraystretch}{0.9}
\begin{tabular}{ l | l } 
 \hline
 Category & Description\\
 \hline \hline
 acc & Accusative marker\\
 acomp & Adjectival complement\\
 acomp-lvc & Complement Light verb construction\\
 advcl & Adverbial clause modifier\\
 advmod & Adverbial modifier\\
 amod & Adjectival modifier\\
 appos & Appositional modifier\\
 aux & Auxiliary\\
 auxpass & Passive auxiliary\\
 cc & Coordination\\
 ccomp & Clausal complement\\
 complm & Complementizer\\
 conj & Conjunct\\
 cop & Copula\\
 cpobj & Object of comparative\\
 cprep & Comparative modifier\\
 dep & Dependent\\
 dep-top & Topic Dependent\\
 dep-voc & Vocative Dependent\\
 det & Determiner\\
 dobj & Direct object\\
 dobj-lvc & Object Light verb construction\\
 fw & foreign word\\
 mark & Marker\\
 mwe & Multi-word expression\\
 neg & Negation modifier\\
 nn & Noun compound modifier\\
 npadvmod & Nominal adverbial modifier\\
 nsubj & Nominal subject\\
 nsubj-lvc & Subject Light verb construction\\
 nsubjpass & Passive nominal subject\\
 num & Numeric modifier\\
 number & Element of compound number\\
 parataxis & Parataxis\\
 pobj & Object of a preposition\\
 poss & Possession modifier\\
 predet & Predeterminer\\
 prep & Prepositional modifier\\
 prep-lvc & Modifier Light verb construction\\
 prt & Phrasal verb particle\\
 punct & Punctuation\\
 quantmod & Quantifier phrase modifier\\
 rcmod & Relative clause modifier\\
 rel & Relative\\
 root & Root\\
 tmod & Temporal modifier\\
 xcomp & Open clausal complement\\
 \hline
\end{tabular}}\end{center}
\caption{Syntactic relations in UPDT.}
\label{table:SRUPDT}
\end{table}

\subsubsection{Colloquialism Temporal Network}
In this section, we study the Deep Net for colloquial language understanding and then describe our new proposed ``Colloquialism Temporal Network" (CATNet) for multi-task word classification and prediction. This network is producing the final translation by ensembling decoding output of the translation models. The embedding layer is a matrix of shape $(v + 1)$ where $v$ is the vocabulary size and $d$ is the dimension of each word vector. We set $v$ to 10M, $d$ to 400, and choose weights at random from $(0, 0.01)$. As for the word embedding model, we use GloVe \cite{pennington2014glove}. Therefore, we feed corresponding matching words in $v$ with its high-level annotation labels. CATNet enhance temporal convolution with a kernel with zero padding, relu activation function, and global max pooling over time to reduce dimensionality. Furthermore, sigmoid activation function is also used, and dropout layers are utilized to regularize the network and prevent overfitting. Using the embedding layer as in CNN models, we implement RNN models using the LSTM layers. Precisely, we implemented a random-embedding model as well as a GloVe embedding model. The word embedding dimension is set to 64 and 400 for both models, respectively. Experiments show that Adam optimizer outperforms SGD. We trained the models using Mean Squared Error (MSE) as well as Binary Cross Entropy loss (BCE). BCE outperformed MSE as expected for polarity classification. Using the same number of LSTM units, dense units and loss function, the GloVe-based model slightly outperformed the random-based one. The RNN model outperformed the CNN model, as anticipated, for text data. The proposed method has been implemented with the TensorFlow \cite{abadi2016tensorflow} and ran on a machine with GeForce RTX 2080 Ti. In the future work, we plan to encode first n-words in translation results to further improve the quality.

\section{Conclusion}
This work presents the ``Large Scale Colloquial Persian Dataset" (LSCP), a large-scale multi-purpose language understanding dataset with comprehensive tasks and annotations. It contains 27M tweets in total with 120M sentences, which is richly labeled over 64M labels encompassing universal and treebank-specific POS tags with dependency relation. We believe that the LSCP dataset as an important source to learn informal Persian language representations which will enable many real-world applications. For the future plan, we are going to expand the dataset with similar rich semantic labels and also provide annotations for other important tasks.

\section{Acknowledgements}
The research was supported by the grants 19-26934X (NEUREM3) of the Czech Science Foundation. The authors would like to thank Kinal Mehta, for his invaluable help and cooperation in this project.

\hfill 
\section{Bibliographical References}
\label{main:ref}

\bibliographystyle{lrec}
\bibliography{LSCP}

\begin{thebibliography}{}

\bibitem[\protect\citename{Abadi \bgroup et al.\egroup
  }2016]{abadi2016tensorflow}
Abadi, M., Barham, P., Chen, J., Chen, Z., Davis, A., Dean, J., Devin, M.,
  Ghemawat, S., Irving, G., Isard, M., et~al.
\newblock (2016).
\newblock Tensorflow: A system for large-scale machine learning.
\newblock In {\em 12th Symposium on Operating Systems Design and
  Implementation}, pages 265--283.

\bibitem[\protect\citename{{Abdi Khojasteh} \bgroup et al.\egroup
  }2020a]{lindatData}
{Abdi Khojasteh}, H., Ansari, E., and Bohlouli, M.
\newblock (2020a).
\newblock Large-scale colloquial persian 0.5.
\newblock {LINDAT}/{CLARIAH}-{CZ} digital library at the Institute of Formal
  and Applied Linguistics ({{\'U}FAL}), Faculty of Mathematics and Physics,
  Charles University, https://hdl.handle.net/11234/1-3195.

\bibitem[\protect\citename{{Abdi Khojasteh} \bgroup et al.\egroup
  }2020b]{khojasteh2020deep}
{Abdi Khojasteh}, H., Ansari, E., Razzaghi, P., and Karimi, A.
\newblock (2020b).
\newblock Deep multimodal image-text embeddings for automatic cross-media
  retrieval.
\newblock {\em arXiv preprint arXiv:2002.10016}.

\bibitem[\protect\citename{AleAhmad \bgroup et al.\egroup
  }2009]{aleahmad2009hamshahri}
AleAhmad, A., Amiri, H., Darrudi, E., Rahgozar, M., and Oroumchian, F.
\newblock (2009).
\newblock Hamshahri: A standard persian text collection.
\newblock {\em Knowledge-Based Systems}, 22(5):382--387.

\bibitem[\protect\citename{Ansari \bgroup et al.\egroup
  }2019]{ansari2019persian}
Ansari, E., {\v{Z}}abokrtsk{\`y}, Z., Haghdoost, H., and Nikravesh, M.
\newblock (2019).
\newblock Persian morphologically segmented lexicon 0.5. lindat/clarin digital
  library at the institute of formal and applied linguistics ({\'u}fal),
  faculty of mathematics and physics, charles university.

\bibitem[\protect\citename{Barari and QasemiZadeh}2005]{barari2005clonizer}
Barari, L. and QasemiZadeh, B.
\newblock (2005).
\newblock Clonizer spell checker adaptive language independent spell checker.
\newblock In {\em AIML 2005 Conference CICC, Cairo, Egypt}, pages 19--21.

\bibitem[\protect\citename{Bijankhan}2004]{bijankhan2004role}
Bijankhan, M.
\newblock (2004).
\newblock The role of the corpus in writing a grammar: An introduction to a
  software.
\newblock {\em Iranian Journal of Linguistics}, 19(2):48--67.

\bibitem[\protect\citename{Conneau \bgroup et al.\egroup
  }2018]{conneau2018xnli}
Conneau, A., Rinott, R., Lample, G., Williams, A., Bowman, S.~R., Schwenk, H.,
  and Stoyanov, V.
\newblock (2018).
\newblock Xnli: Evaluating cross-lingual sentence representations.
\newblock In {\em Proceedings of the 2018 Conference on Empirical Methods in
  Natural Language Processing}. Association for Computational Linguistics.

\bibitem[\protect\citename{Darrudi \bgroup et al.\egroup
  }2008]{darrudi2008dorir}
Darrudi, E., Baradaran~Hashemi, H., AleAhmad, A., Zare~Bidoki, A., Habibian,
  A., Mahdikhani, F., Shakery, A., and Rahgozar, M.
\newblock (2008).
\newblock dorir collection for persian web retrieval.
\newblock Technical report, Technical Report No. DBRG-TR-02.

\bibitem[\protect\citename{Erjavec}2012]{erjavec2012multext}
Erjavec, T.
\newblock (2012).
\newblock Multext-east: morphosyntactic resources for central and eastern
  european languages.
\newblock {\em Language resources and evaluation}, 46(1):131--142.

\bibitem[\protect\citename{Hafezi and Rezaeian}2018]{hafezi2018neural}
Hafezi, L. and Rezaeian, M.
\newblock (2018).
\newblock Neural architecture for persian named entity recognition.
\newblock In {\em 2018 4th Iranian Conference on Signal Processing and
  Intelligent Systems (ICSPIS)}, pages 61--64. IEEE.

\bibitem[\protect\citename{Karimi \bgroup et al.\egroup
  }2017]{karimi2017extracting}
Karimi, A., Ansari, E., and Bigham, B.~S.
\newblock (2017).
\newblock Extracting an english-persian parallel corpus from comparable
  corpora.
\newblock {\em arXiv preprint arXiv:1711.00681}.

\bibitem[\protect\citename{Klein \bgroup et al.\egroup }2017]{klein2017opennmt}
Klein, G., Kim, Y., Deng, Y., Senellart, J., and Rush, A.~M.
\newblock (2017).
\newblock Opennmt: Open-source toolkit for neural machine translation.
\newblock In {\em Proceedings of ACL 2017, System Demonstrations}, pages
  67--72.

\bibitem[\protect\citename{Koehn \bgroup et al.\egroup }2007]{koehn2007moses}
Koehn, P., Hoang, H., Birch, A., Callison-Burch, C., Federico, M., Bertoldi,
  N., Cowan, B., Shen, W., Moran, C., Zens, R., et~al.
\newblock (2007).
\newblock Moses: Open source toolkit for statistical machine translation.
\newblock In {\em Proceedings of the 45th annual meeting of the association for
  computational linguistics companion volume proceedings of the demo and poster
  sessions}, pages 177--180.

\bibitem[\protect\citename{Lo \bgroup et al.\egroup }2017]{lo2017multilingual}
Lo, S.~L., Cambria, E., Chiong, R., and Cornforth, D.
\newblock (2017).
\newblock Multilingual sentiment analysis: from formal to informal and scarce
  resource languages.
\newblock {\em Artificial Intelligence Review}, 48(4):499--527.

\bibitem[\protect\citename{Majli{\v{s}}}2011]{majlivs2011w2c}
Majli{\v{s}}, M.
\newblock (2011).
\newblock W2c--web to corpus--corpora.

\bibitem[\protect\citename{Nivre \bgroup et al.\egroup
  }2017]{nivre2017universal}
Nivre, J., Agi{\'c}, {\v{Z}}., Ahrenberg, L., et~al.
\newblock (2017).
\newblock Universal dependencies 2.0. lindat/clarin digital library at the
  institute of formal and applied linguistics, charles university, prague.

\bibitem[\protect\citename{Pennington \bgroup et al.\egroup
  }2014]{pennington2014glove}
Pennington, J., Socher, R., and Manning, C.
\newblock (2014).
\newblock Glove: Global vectors for word representation.
\newblock In {\em Proceedings of the 2014 conference on empirical methods in
  natural language processing (EMNLP)}, pages 1532--1543.

\bibitem[\protect\citename{Pilevar \bgroup et al.\egroup }2011]{pilevar2011tep}
Pilevar, M.~T., Faili, H., and Pilevar, A.~H.
\newblock (2011).
\newblock Tep: Tehran english-persian parallel corpus.
\newblock In {\em International Conference on Intelligent Text Processing and
  Computational Linguistics}, pages 68--79. Springer.

\bibitem[\protect\citename{Qi \bgroup et al.\egroup }2018]{qi2018universal}
Qi, P., Dozat, T., Zhang, Y., and Manning, C.~D.
\newblock (2018).
\newblock Universal dependency parsing from scratch.
\newblock In {\em Proceedings of the {CoNLL} 2018 Shared Task: Multilingual
  Parsing from Raw Text to Universal Dependencies}, pages 160--170, Brussels,
  Belgium, October. Association for Computational Linguistics.

\bibitem[\protect\citename{Rasooli \bgroup et al.\egroup
  }2013]{rasooli2013development}
Rasooli, M.~S., Kouhestani, M., and Moloodi, A.
\newblock (2013).
\newblock Development of a persian syntactic dependency treebank.
\newblock In {\em Proceedings of the 2013 Conference of the North American
  Chapter of the Association for Computational Linguistics: Human Language
  Technologies}, pages 306--314.

\bibitem[\protect\citename{Seraji}2015]{seraji2015morphosyntactic}
Seraji, M.
\newblock (2015).
\newblock {\em Morphosyntactic Corpora and Tools for Persian}.
\newblock {Ph.D.} thesis, Acta Universitatis Upsaliensis.

\bibitem[\protect\citename{Shamsfard \bgroup et al.\egroup
  }2010]{shamsfard2010semi}
Shamsfard, M., Hesabi, A., Fadaei, H., Mansoory, N., Famian, A., Bagherbeigi,
  S., Fekri, E., Monshizadeh, M., and Assi, S.~M.
\newblock (2010).
\newblock Semi automatic development of farsnet; the persian wordnet.
\newblock In {\em Proceedings of 5th global WordNet conference, Mumbai, India},
  volume~29.

\bibitem[\protect\citename{Tiedemann}2012]{tiedemann2012parallel}
Tiedemann, J.
\newblock (2012).
\newblock Parallel data, tools and interfaces in opus.
\newblock In Nicoletta Calzolari~(Conference Chair), et~al., editors, {\em
  Proceedings of the Eight International Conference on Language Resources and
  Evaluation (LREC'12)}, volume 2012, pages 2214--2218. European Language
  Resources Association (ELRA).

\bibitem[\protect\citename{Vaswani \bgroup et al.\egroup
  }2017]{vaswani2017attention}
Vaswani, A., Shazeer, N., Parmar, N., Uszkoreit, J., Jones, L., Gomez, A.~N.,
  Kaiser, {\L}., and Polosukhin, I.
\newblock (2017).
\newblock Attention is all you need.
\newblock In {\em Advances in neural information processing systems}, pages
  5998--6008.

\bibitem[\protect\citename{Zhao and Ji}2018]{zhao2018lsicc}
Zhao, J. and Ji, Z.
\newblock (2018).
\newblock Lsicc: A large scale informal chinese corpus.
\newblock {\em arXiv preprint arXiv:1811.10167}.

\end{thebibliography}

\end{document}